\documentclass[11pt,a4paper]{article}
\usepackage[hyperref]{emnlp2020}
\usepackage{times}
\usepackage{bbm}
\usepackage{amsmath}
\usepackage{amsfonts}
\usepackage{microtype}
\usepackage{graphicx}
\usepackage{caption}
\usepackage{subcaption}
\usepackage{subfiles}
\usepackage{latexsym}
\usepackage{algorithmic}
\usepackage[ruled]{algorithm2e}

\DeclareMathOperator*{\argmax}{argmax}

\usepackage{microtype}

\aclfinalcopy 
\def\aclpaperid{128} 

\title{Structured Attention for Unsupervised Dialogue Structure Induction}

\author{Liang Qiu{\normalfont\textsuperscript{1}}, Yizhou Zhao{\normalfont\textsuperscript{1}}, Weiyan Shi{\normalfont\textsuperscript{2}}, Yuan Liang{\normalfont\textsuperscript{3}}, Feng Shi{\normalfont\textsuperscript{1}}, \\
{\bf Tao Yuan{\normalfont\textsuperscript{1}}, Zhou Yu{\normalfont\textsuperscript{2}}, Song-Chun Zhu{\normalfont\textsuperscript{1}}} \\
\textsuperscript{1}UCLA Center for Vision, Cognition, Learning, and Autonomy \\
\textsuperscript{2}University of California, Davis \\
\textsuperscript{3}University of California, Los Angeles \\
\texttt{\{liangqiu,yizhouzhao,liangyuandg,shifeng,taoyuan\}@ucla.edu} \\
\texttt{\{wyshi,joyu\}@ucdavis.edu} \\
\texttt{sczhu@stat.ucla.edu}}

\date{}

\begin{document}
\maketitle
\begin{abstract}
Inducing a meaningful structural representation from one or a set of dialogues is a crucial but challenging task in computational linguistics. Advancement made in this area is critical for dialogue system design and discourse analysis. It can also be extended to solve grammatical inference. In this work, we propose to incorporate structured attention layers into a Variational Recurrent Neural Network (VRNN) model with discrete latent states to learn dialogue structure in an unsupervised fashion. Compared to a vanilla VRNN, structured attention enables a model to focus on different parts of the source sentence embeddings while enforcing a structural inductive bias. Experiments show that on two-party dialogue datasets, VRNN with structured attention learns semantic structures that are similar to templates used to generate this dialogue corpus. While on multi-party dialogue datasets, our model learns an interactive structure demonstrating its capability of distinguishing speakers or addresses, automatically disentangling dialogues without explicit human annotation.\footnote{The code is released at \url{https://github.com/Liang-Qiu/SVRNN-dialogues}.}
\end{abstract}

\section{Introduction}
\begin{figure}[ht]
\vskip 0.2in
\begin{center}
\centerline{\includegraphics[width=\linewidth]{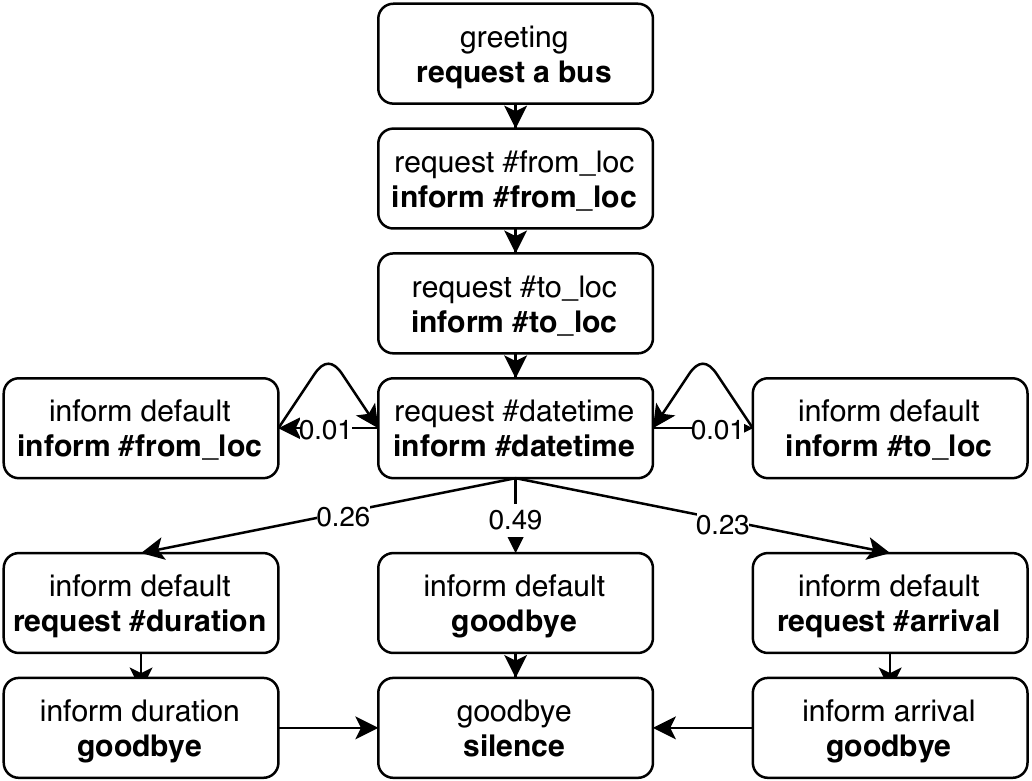}}
\caption{Original dialogue structure of the bus information request domain in SimDial \cite{zhao-eskenazi-2018-zero}. User intents are marked in bold.}
\label{fig:bus}
\end{center}
\vskip -0.2in
\end{figure}

\begin{figure}[ht]
\vskip 0.2in
\begin{center}
\centerline{\includegraphics[width=\linewidth]{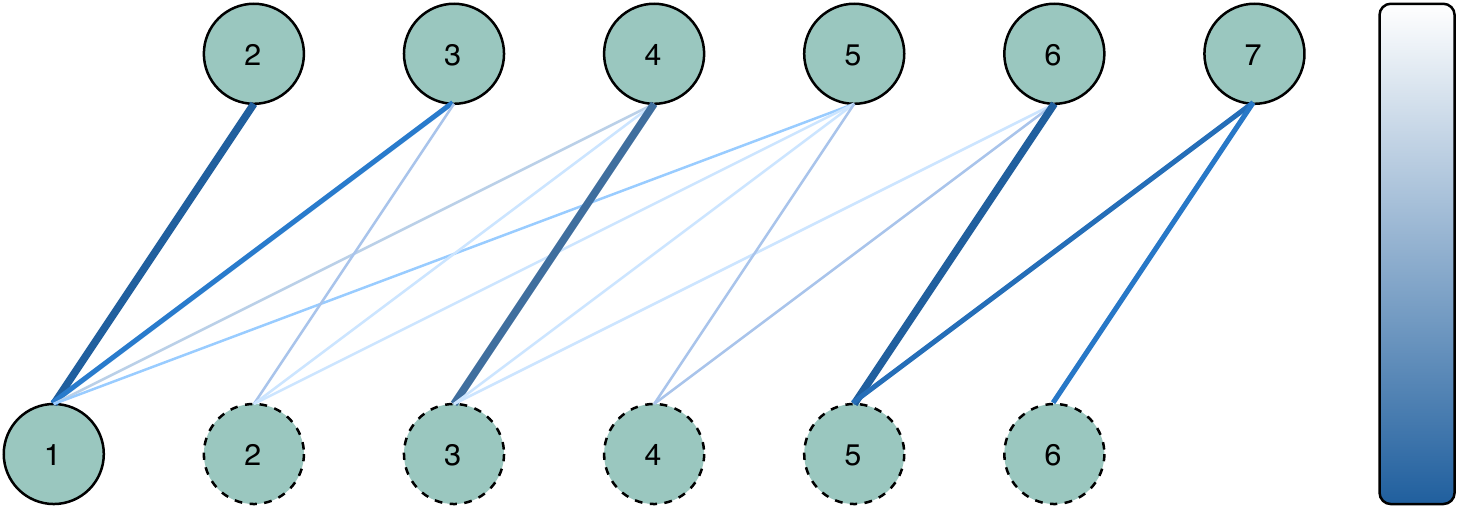}}
\caption{Learned interactive structure from a multi-party dialogue sample in Ubuntu Chat Corpus \cite{uthus2013ubuntu}.}
\label{fig:visualization}
\end{center}
\vskip -0.2in
\end{figure}

Grammatical induction for capturing a structural representation of knowledge has been studied for some time \cite{de2010grammatical}. Given the achievement in related areas like learning \textit{Hidden Markov} acoustic models in speech recognition \cite{bahl1986maximum} and sentence dependency parsing in language understanding \cite{covington2001fundamental}, our work aims to explore a more sophisticated topic: learning structures in dialogues. Figure \ref{fig:bus} shows the underlying semantic structure of conversations about bus information request from SimDial dataset \cite{zhao-eskenazi-2018-zero}, with one example dialogue as shown in Table \ref{tab:simdial-example}. Another interesting type of dialogue structure is the interactive structure in multi-party dialogues. Figure \ref{fig:visualization} illustrates the interactive structure we learned from a dialogue sample in Ubuntu Chat Corpus \cite{lowe-etal-2015-ubuntu}. Each node represents an utterance from different speakers in the dialogue with darker linkages represent stronger dependency relations between utterances. When speaker/addressee information is unavailable in the corpus, learning such a structure allows disentangling the conversation \cite{serban2015text} and estimating the speaker labels. Discovering dialogue structures is crucial for various areas in computational linguistics, such as dialogue system building \cite{young2006using}, discourse analysis \cite{grosz-sidner-1986-attention}, and dialogue summarization \cite{murray2005extractive, liu-etal-2010-dialogue}. Through looking into this topic, we can further improve the capability of machines to learn more generalized, interpretable knowledge representation from data. 

However, capturing structure from the conversation is still much under-explored. The complexity of dialogues could range from several-round task-oriented dialogues to tens-round multi-party chitchat. It is unclear that for these different categories of dialogues, what types of inductive biases or constraints we could add to reduce the search space. It also remains an unsolved question for formally evaluating the performance of dialogue structure induction algorithms. In this paper, we propose to use a combination of structured attention and unsupervised generative model to infer the latent structure in a dialogue.

Specifically, instead of simply applying a softmax function on potentials between a decoder query and encoder hidden states, dynamic programming algorithms like \textit{Forward-Backward} \cite{devijver1985baum} and \textit{Inside-Outside} \cite{lari1990estimation} could be used to efficiently calculate marginal probabilities from pairwise potentials with a structural constraint. Through embedding such structured attention layers in a \textit{Variational Recurrent Neural Network} (VRNN) model, we can learn latent structures in dialogues by jointly re-generating training dialogues. Such a process requires no human annotation and is useful for dialogue analysis. In addition, by selecting appropriate structural biases or constraints, we can learn not only semantic structures but also interactive structures. A linear \textit{Conditional Random Field} (CRF) attention layer is used in two-party dialogues to discover semantic structures. A non-projective dependency tree attention layer is embedded to learn an interactive structure that could help identify speaker/addressee information in multi-party dialogues that have tangled conversation threads, such as forum discussions. 

This paper makes the following contributions. We propose to incorporate a structured attention layer in VRNN to learn latent structures in dialogues. To our knowledge, no work connecting structured attention with unsupervised dialogue structure learning has been done. We prove our proposed VRNN-LinearCRF learns better structures than the baseline VRNN on the SimDial dataset for semantic structure learning in two-party dialogues. For interactive structure learning in multi-party dialogues, we combine VRNN with a non-projective dependency tree attention layer. It achieves similar generation performance as the baseline GSN model \cite{hu2019gsn} on Ubuntu Chat Corpus \cite{uthus2013ubuntu, lowe-etal-2015-ubuntu}, while our model can identify the speaker/addressee information without trained on explicit labels. We release our code as well as the processed datasets to help stimulate related researches.

\section{Related Work}
Attention mechanism \cite{vaswani2017attention} has been widely adopted as a way for embedding categorical inference in neural networks for performance gain and interpretability \cite{jain-wallace-2019-attention, wiegreffe-pinter-2019-attention}. However, for many tasks, we want to model richer structural dependencies without abandoning end-to-end training. \textit{Structured Attention Networks} \cite{kim2017structured} can extend attention beyond the standard soft-selection approach by attending to partial segments or subtrees. People have proven its effectiveness on a variety of synthetic and real tasks: tree transduction, neural machine translation, question answering, and natural language inference \cite{alex2020torchstruct}. In this paper, we propose to utilize structured attention to explore dialogue structures. Specifically, we work on two types of dialogue structures,  semantic structures (dialogue intent transitions), and interactive structures (addressee/speaker changes). 

Semantic structures have been studied extensively. Some previous works, such as \cite{jurafsky1997switchboard}, learned semantic structures relying on human annotations, while such annotations are costly and can vary in quality. Other unsupervised studies used \textit{Hidden Markov Model} (HMM) \cite{chotimongkol2008learning, ritter-etal-2010-unsupervised, zhai-williams-2014-discovering}. Recently, \textit{Variational Autoencoders} (VAEs) \cite{kingma2013auto} and their recurrent version, \textit{Variational Recurrent Neural Networks} (VRNNs) \cite{chung2015recurrent}, connects neural networks and traditional Bayes methods. Because VRNNs apply a point-wise non-linearity to the output at every timestamp, they are also more suitable to model highly non-linear dynamics over the simpler dynamic Bayesian network models. \citet{serban2017hierarchical} proposed the VHRED model by combining the idea of VRNNs and \textit{Hierarchical Recurrent Encoder-Decoder} (HRED) \cite{sordoni2015hierarchical} for dialogue generation. Similarly, \citet{zhao-etal-2018-unsupervised} proposed to use VAEs to learn discrete sentence representations. \citet{shi-etal-2019-unsupervised} used two variants of VRNNs to learn the dialogue semantic structures and discussed how to use learned structure to improve reinforcement learning-based dialogue systems. But none of the previous work has tried to incorporate structured attention in VRNNs to learn dialogue structure.

Compared to semantic structures, the interactive structure of dialogues is not clearly defined. \citet{elsner-charniak-2008-talking} initiated some work about dialogue disentanglement, which is defined as dividing a transcript into a set of distinct conversations. \citet{serban2015text} tested standard RNN and its conditional variant for turn taking and speaker identification. Both of the tasks are highly related to understanding the interactive structure but not identical. Our task, different from both of them, aims to construct an utterance dependency tree to represent a multi-party dialogue's turn taking. The tree can not only be used to disentangle the conversations but also label each utterance's speakers and addressees. We compare our model with \textit{Graph Structured Network} (GSN), recently proposed by \citet{hu2019gsn}. GSN builds a conversation graph utilizing explicit speaker/addressee information in Ubuntu Chat Corpus \cite{uthus2013ubuntu} to improve the dialogue generation performance. Our model shows similar generation performance as them while demonstrating its capability of learning the utterance dependency tree.

\section{Problem Formulations}
We discuss the semantic and interactive dialogue structure learning separately. In task-oriented two-party dialogues (between system and user), we want to discover a semantic probabilistic grammar shared by dialogues in the same domain. While for multi-party dialogues, \textit{e.g.}, conversations in a chatroom, which may have multiple conversations occur simultaneously, we are more interested in finding an interactive structure that could help disentangle the conversation and identify the speakers/addressees. Our method of structure learning is flexible to handle both problems with the formulations as shown below.

For semantic dialogue structure learning, we formulate the problem as labeling the dialogue with a sequence of latent states. Each conversational exchange $x_i$ (a pair of system and user utterances at time step $i$) belongs to a latent state $z_i$, which has an effect on the future latent states and the words the interlocutors produce. The latent dialogue state is defined to be discrete, \textit{i.e.}, $z_i \in \{1, 2, ..., N\}$, where $N$ is the number of states predefined from experience. Our goal is to generate the current sentence pair $x_i$ that maximizes the conditional likelihood of $x_i$ given the dialogue history while jointly learning a latent state sequence $\textbf{z} = [z_1, z_ 2,..., z_n]$:
\begin{equation}
    \mathbf{\hat{x}} =\argmax_\mathbf{x}\sum_{i=1}^{|\mathbf{x}|}\log(P(\mathbf{z}_{< i}|\mathbf{x}_{< i})P(x_i|\mathbf{z}_{< i})).
\end{equation} 
Then, we can induce a probabilistic dialogue grammar by estimating the state transition probabilities through maximizing the likelihood of the parsed latent state sequences.
\begin{figure*}[ht]
\vskip 0.2in
\begin{center}
\centerline{\includegraphics[width=\linewidth]{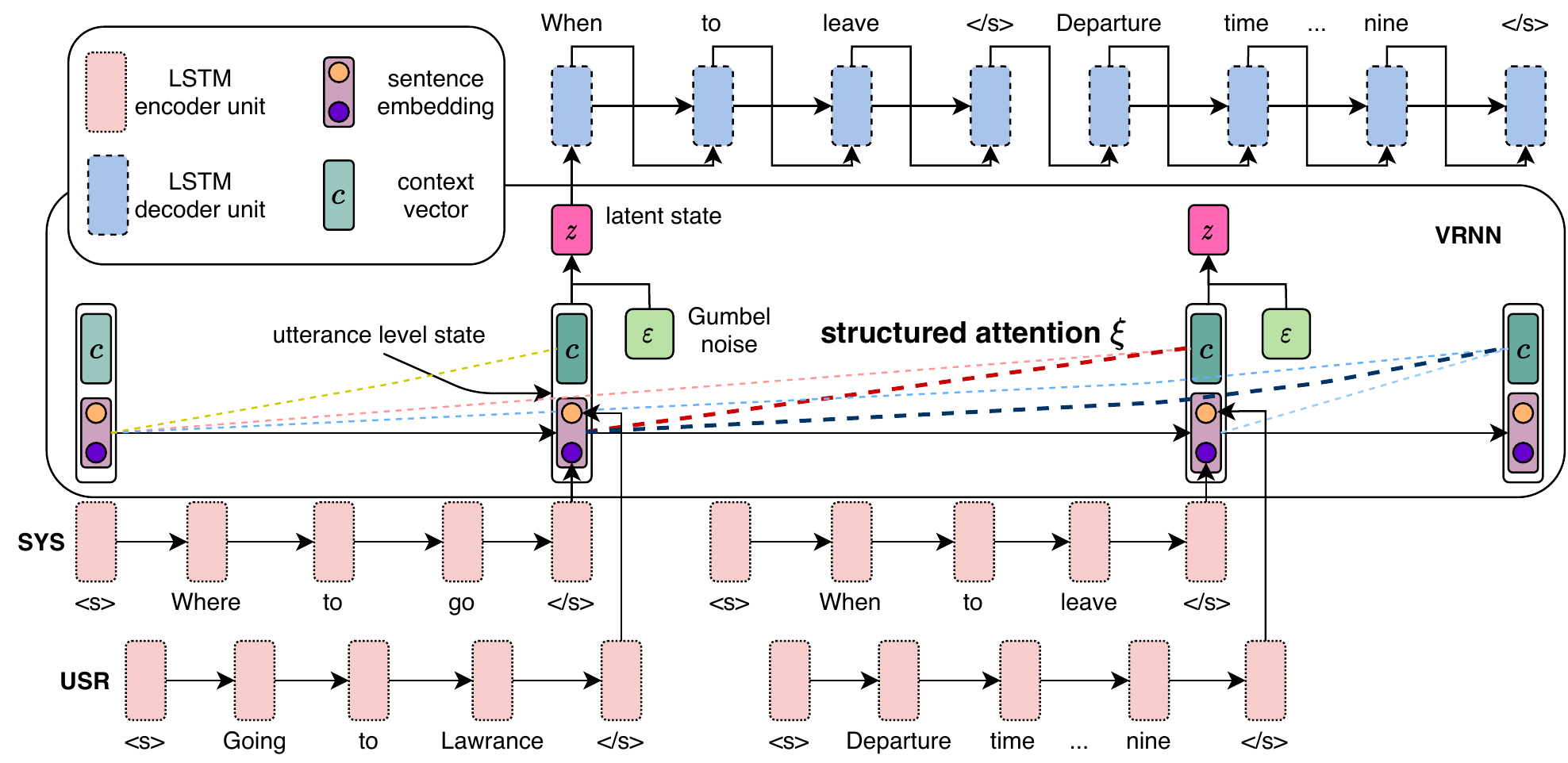}}
\caption{Structured-Attention Variational Recurrent Neural Network (SVRNN)}
\label{tsan}
\end{center}
\vskip -0.2in
\end{figure*}

A multi-party dialogue session can be formulated as an utterance-level dependency tree $\mathbf{T}(V,E)$, where $V$ is the set of nodes encoding the utterances, $E=\{e_{i,j}\}_{i<j}^m \in \{0, 1\}$ indicates whether utterance $i$ is the parent of utterance $j$, and $m$ is the maximum number of possible edges. 
\begin{equation}
    \begin{aligned}
        \mathbf{\hat{x}} &=\argmax_\mathbf{x}\sum_{i=1}^{|\mathbf{x}|}\log(P(\mathbf{T}|\mathbf{x}_{<i})P(x_i|\mathbf{T}))    \\
                              &=\argmax_\mathbf{x}\sum_{i=1}^{|\mathbf{x}|}\log(\prod_{j<k}^{i-1} P(e_{j,k}=1|\mathbf{x}_{<i}) \cdot \\
                              &\quad\quad\quad\quad\quad\quad\quad\quad P(x_i|\mathbf{T}))   \\
                              &=\argmax_\mathbf{x}\Big[\sum_{i=1}^{|\mathbf{x}|}\sum_{j<k}^{i-1}\log( P(e_{j,k}=1|\mathbf{x}_{<i}) \, + \\
                              &\quad\quad\quad\quad\quad\sum_{i=1}^{|\mathbf{x}|}\log (P(x_i|\mathbf{T})\Big]
    \end{aligned}
    \label{eq:2}
\end{equation} 
Each path of the dependency tree represents a thread in the multi-party conversation in chronological order. Our goal is to generate the response $\mathbf{\hat{x}}$ that maximizes the conditional likelihood of the response given the dialogue history while jointly learning a latent utterance dependency tree as shown in Equation \ref{eq:2}. The conditional likelihood is factorized into two parts, representing the encoding and decoding processes respectively. We can further reason about the speaker/addressee labels or disentangle the conversation by clustering the utterances from the learned tree.

\section{Variational Recurrent Neural Network with Structured Attention}
The overall architecture of Structured-Attention Variational Recurrent Neural Network (SVRNN) is illustrated in Figure \ref{tsan}. The LSTM \cite{hochreiter2001long} word-level encoder marked in pink encodes each utterance into a sentence embedding. Then an utterance-level encoder VRNN with different structured attention layers encodes the dialogue history into a latent state $z$. A decoder marked in blue will decode the next utterances from the latent state. We describe more details about the key components of our model in the following subsections.

\subsection{Variational Recurrent Neural Network}
The pursuit of using an autoencoder like \textit{Variational Recurrent Neural Network} (VRNN) is to compress the essential information of the dialogue history into a lower-dimensional latent code. The latent code $z$ is a random vector sampled from a prior $p(z)$ and the data generation model is described by $p(x|z)$. The VRNN contains a \textit{Variational Autoencoder} (VAE) at each time step. The VAE consists of an encoder $q_\lambda(z|x)$ for approximating the posterior $p(z|x)$, and a decoder $p_\theta(x|z)$ for representing the distribution $p(x|z)$. The variational inference attains its maximum likelihood by maximizing \textit{evidence lower bound} (ELBO):
\begin{equation}
\mathbb{E}\left[\log p_{\theta}(x | z)\right]-\operatorname{KL}\left(q_{\lambda}(z | x) \| p(z)\right) \leq \log p(x).
\end{equation}

For sequential data, the parameterization of the generative model is factorized by the posterior $ p\left(z_{t} | x_{<t}, z_{<t}\right)$ and the generative model $p\left(x_{t} | z_{\leq t}, x_{<t}\right)$, \textit{i.e.},
\begin{equation}
    \begin{aligned}
        p(x \leq T, z \leq T) = \prod_{t=1}^{T} & \big[ p \left(x_{t} | z_{\leq t}, x_{<t}\right) \cdot \\
        & p\left(z_{t} | x_{<t}, z_{<t}\right) \big].
    \end{aligned}
    \label{eq:generative}
\end{equation}
The learning objective function becomes maximizing the ELBO for all time steps
\begin{equation}
\begin{split}
\mathbb{E}\Big[\sum_{t=1}^{T}(&-\mathrm{KL}\left(q(z_{t} | x_{\leq t}, z_{<t}) \| p(z_{t} | x_{<t}, z_{<t})\right)\\
&+\log p(x_{t} | z_{\leq t}, x_{<t}))\Big].
\end{split}
\end{equation}
In addition, to mitigate the \textit{vanishing latent variable problem} in VAE, we incorporate Bag-of-Words (BOW) loss and Batch Prior Regularization (BPR) \cite{zhao-etal-2017-learning} with a tunable weight $\lambda$. By adjusting the $\lambda$, the VRNN based models can achieve a balance between clustering the utterance surface formats and attention on the context.

\subsection{Linear CRF Attention}
As we formulate the semantic structure learning in two-party dialogues as a state tagging problem, we find it suitable to use a linear-chain \textit{Conditional Random Field} (CRF) attention layer with VRNN. Define $\boldsymbol{\xi}$ to be a random vector $\boldsymbol{\xi}=[\xi_1,...,\xi_n]$ with $\xi_i\in \{0,1\}$. $n$ is the number of utterances in a dialogue. The context vector $\mathbf{c}_j$ given the current sentence hidden state $\mathbf{h}_j$ and hidden state history $\mathbf{h}$ can thus be written as:
\begin{equation}
    \textbf{c}_j = \sum_{i=1}^{j-1} p(\xi_i=1|\textbf{h},\textbf{h}_j)\textbf{h}_i.
\end{equation}
We model the distribution over the latent variable $\boldsymbol{\xi}$ with a linear-chain CRF with pairwise edges,
\begin{equation}
    p(\xi_1,...,\xi_n|\textbf{h},\textbf{h}_j) = softmax(\sum_{i=1}^{j-2}\theta_{i, i+1}(\xi_i,\xi_{i+1})),
\end{equation}
where $\theta_{i,i+1}(k,l)$ is the pairwise potential for $\xi_i=k$ and $\xi_{i+1}=l$. The attention layer is a two-state CRF where the unary potentials at the $j$-th dialogue turn are:
\begin{equation}
    \theta_i(k)=\begin{cases}
                \textbf{h}_i\textbf{W}_1\textbf{h}_j, k=0 \\
                \textbf{h}_i\textbf{W}_2\textbf{h}_j, k=1
                \end{cases},
\end{equation}
where [$\textbf{h}_1,...,\textbf{h}_n$] are utterance level hidden states and $\textbf{W}_1,\textbf{W}_2$ are parameters. The pairwise potentials can be parameterized as
\begin{equation}
    \theta_{i,i+1}(\xi_i,\xi_{i+1})=\theta_i(\xi_i)+\theta_{i+1}(\xi_{i+1})+\textbf{h}_i^\top\textbf{h}_{i+1}.
\end{equation}

The marginal distribution $p(\xi_i=1|x)$ can be calculated efficiently in linear-time for all $i$ using message-passing, \textit{i.e.}, the \textit{forward-backward} shown in Algorithm \ref{alg:forward-backward}.

\begin{algorithm}
    \caption{Forward-Backward for LinearCRF Attention}
    \label{alg:forward-backward}
\begin{algorithmic}
  \STATE {\bfseries Input:} potential $\theta$
  \STATE $\alpha[0,\langle t \rangle] \leftarrow 0$
  \STATE $\beta[n+1,\langle t \rangle] \leftarrow 0$
  \FOR{$i=1,...,n;c\in \mathcal{C}$}
        \STATE $\alpha[i,c] \leftarrow \bigoplus_y \alpha[i-1,y] \otimes \theta_{i-1,i}[y,c]$
   \ENDFOR
   \FOR{$i=n,...,1;c\in \mathcal{C}$}
        \STATE $\beta[i,c] \leftarrow \bigoplus_y \beta[i+1,y] \otimes \theta_{i,i+1}[c,y]$
   \ENDFOR
   \STATE $A \leftarrow \alpha[n+1,\langle t \rangle]$
   \FOR{$i=1,...,n;c\in \mathcal{C}$}
        \STATE $p(\xi_i=c|x) \leftarrow \exp(\alpha[i,c]\otimes\beta[i,c]\otimes-A)$
   \ENDFOR
   \STATE \Return $p$
\end{algorithmic}
\end{algorithm}
$\mathcal{C}$ denotes the state space and $\langle t \rangle$ is the special start/stop state. Typically the forward-backward with marginals is performed in the log-space semifield $\mathbb{R} \cup \{\pm \infty\}$ with binary operations $\oplus=$ logadd and $\otimes=+$ for numerical precision. These marginals allow us to calculate the context vector. Crucially, the process from vector softmax to \textit{forward-backward} algorithm is a series of differentiable steps, and we can compute the gradient of the marginals with respect to the potentials \cite{kim2017structured}. This allows the linear CRF attention layer to be trained end-to-end as a part of the VRNN. 

\subsection{Non-projective Dependency Tree Attention}
For interactive structure learning in multi-party dialogues, we want to learn an utterance dependency tree from each dialogue. Therefore, we propose to use a non-projective dependency tree attention layer with VRNN for this purpose. The potentials $\theta_{i,j}$, which reflect the score of selecting the $i$-th sentence being the parent of the $j$-th sentence (\textit{i.e.}, $x_i\rightarrow x_j$), can be calculated by
\begin{equation}
    \theta_{i,j}=\tanh(\textbf{s}^\top \tanh\textbf{(W}_1\textbf{h}_i + \textbf{W}_2\textbf{h}_j + \textbf{b})),
\end{equation}
where $\textbf{s},\textbf{b},\textbf{W}_1,\textbf{W}_2$ are parameters, $\textbf{h}_i, \textbf{h}_j$ are sentence hidden states.

The probability of a parse tree $\xi$ given the dialogue $x=[x_1,...,x_n]$ is,
\begin{equation}
    \begin{split}
            p(\xi|x)=\textrm{softmax}(&\mathbbm{1}\{\xi\textrm{ is valid}\} \cdot \\
            &\sum_{i\neq j}\mathbbm{1}\{\xi_{i,j}=1\}\theta_{i,j}),
    \end{split}
\end{equation}
where the latent variable $\xi_{i,j}\in\{0,1\}$ for all $i\neq j$ indicates that the $i$-th sentence is the parent of the $j$-th sentence; and $\mathbbm{1}\{\xi\textrm{ is valid}\}$ is a special global constraint that rules out configurations of $\xi_{i,j}$'s that violate parsing constraints. In our case, we specify each sentence has one parent and that must precede the child sentence, i.e,  
\begin{equation}
    \begin{aligned}
    \sum_{i=1}^n\xi_{i,j} = 1    &&    \xi_{i,j} = 0 (i\geq j).
    \end{aligned}
\end{equation}
It is possible to calculate the marginal probability of each edge $p(\xi_{i,j}=1|x)$ for all $i,j$ in $O(n^3)$ time using the \textit{inside-outside} algorithm with details explained in Appendix, which is a generalization of the \textit{forward-backward} algorithm.

Then the soft-parent or the context vector of the $j$-th sentence is calculated using parsing marginals, \textit{i.e.}, 
\begin{equation}
    \textbf{c}_j=\sum_{i=1}^n p(\xi_{i,j}=1|\textbf{h}, \textbf{h}_j)\textbf{h}_i.
\end{equation}
The original embedding is concatenated with its context vector to form the new representation 
\begin{equation}
    \hat{\textbf{h}}_j=[\textbf{h}_j;\textbf{c}_j].
\end{equation}

\subsection{Decoder}
In order to generate a response to an utterance $i$, the decoder calculates a distribution over the vocabulary then sequentially predicts word $w_k$ using a \textit{softmax} function:
\begin{equation}
    \begin{aligned}
        p(\mathbf{w}|\hat{\textbf{h}}) &= \prod_{k=1}^{|\mathbf{w}|}P(w_k|\hat{\textbf{h}},\mathbf{w}_{<k}) \\
                  &=\prod_{k=1}^{|\mathbf{w}|}softmax(MLP(\mathbf{h}_k^{dec},\mathbf{c}_k^{dec}))   \\
        \mathbf{h}_0^{dec} &= \hat{\textbf{h}}_i    \\
        \mathbf{h}_k^{dec} &= LSTM(\mathbf{h}_{k-1}^{dec}, MLP(\mathbf{e}_{w_{k-1}};\mathbf{c}_{k-1}^{dec}))    \\
        \mathbf{c}_k^{dec} &= \sum_{j=1}^i softmax(\mathbf{h}_k^{dec}\mathbf{W}_a\hat{\textbf{h}}_j)\hat{\textbf{h}}_j,
    \end{aligned}
\end{equation}
 where $\hat{\textbf{h}}_i$ is the hidden state for utterance $i$ with structured attention, $\mathbf{h}_k^{dec}$ is the hidden state of the decoder LSTM, $\mathbf{e}_{w_{k-1}}$ is the embedding of the predicted word at decoding time stamp $(k-1)$, and $\mathbf{c}_k^{dec}$ is the attention-based context vector at decoding time stamp $k$. Note that the context vector here is calculated with the simple attention different from the structured attention we described before. $\mathbf{W}_a$ is a matrix to learn the match degree of $\mathbf{h}_k^{dec}$ and $\hat{\textbf{h}}_j$.
 
\section{Experiments}
We incorporate structured attention in VRNNs to explore two types of dialogue structure, semantic structure, and interactive structure.

\subsection{Semantic Structure Learning in Two-party Dialogues}
\subsubsection{Datasets}
We test the VRNN with Linear CRF Attention on the SimDial dataset \cite{zhao-eskenazi-2018-zero} of simulated conversations. Dialogues are generated for information requests in four domains: bus, restaurant, weather, and movie. Table \ref{tab:simdial-example} shows an example dialogue in bus schedule request domain. Despite significant variations exist between dialogues of the same domain, we aim to explore a shared semantic structure among each dialogue domain. We validate our algorithm on this simulated dataset because these dialogues are generated using pre-defined templates that make recovering ground truth structures much easier. One recovered ground truth structure with transition probabilities is shown in Figure \ref{fig:bus}. We have 800 dialogue samples for training, 100 for validation, and 100 for testing in each dialog domain. The length of the dialogues ranges from 6 to 13 utterances. The maximum length of an utterance is 33 words. 
\begin{table}[]
\small
\begin{tabular*}{\linewidth}{cl}
\hline
From & Utterance                      \\ \hline
SYS: & Ask me about bus information. How can I help?\\
USR: & Hi. I need a bus.              \\
SYS: & Where do you want to take off? \\
USR: & Going to Lawrance.             \\
SYS: & What time do you need the bus? \\
USR: & Departure time is 9.           \\
SYS: & Bus 137 can take you there. What else can I do?\\
USR: & Not done yet. How long will it take? \\
SYS: & The ride is 45 minutes long. What else can I do?\\
USR: & No more questions. Thank you.  \\
SYS: & Goodbye.                       \\ \hline
\end{tabular*}
\caption{An example two-party bus information request dialogue in SimDial \cite{zhao-eskenazi-2018-zero}.}
\label{tab:simdial-example}
\end{table}

\subsubsection{Evaluation Metrics}
Since the number of states is unknown during unsupervised training, we set the state number empirically to 10. Then the learned structure is essentially a state transition matrix of size $10\times10$. However, the original structure could be another state transition matrix of any size depending on the domain complexity. This makes the model evaluation on the ground truth a problem because it requires us to measure the difference between two state transition matrices of different sizes. To alleviate this problem, we define two metrics: Structure Euclidean Distance (SED) and Structure Cross-Entropy (SCE). We first estimate a probabilistic mapping $P_{s_i, s'_i}$ between the learned states $\{s'_i, i=1,2,...,M\}$ and the true states $\{s_i, i=1,2,...,N\}$, through dividing the number of utterances that have the ground truth state $s_i$ and learned state $s'_i$ by number of utterances with the ground truth state $s_i$. And we let the reversed mapping probability $P_{s'_i, s_i}$ be the normalized transpose of $P_{s_i, s'_i}$. Then SED and SCE are defined as:
\begin{equation}
    \begin{aligned}
        T'_{s_a, s_b} &= \sum_{i,j \in \{1,2,...,M\}}P_{s_a, s'_i} \cdot T_{s'_i, s'_j} \cdot P_{s'_j, s_b}  \\
        SED &= \frac{1}{N}\sqrt{\sum_{a,b \in \{1,2,...,N\}}(T'_{s_a, s_b}- T_{s_as_b})^2}  \\
        SCE &= \frac{1}{N}\sum_{a,b \in \{1,2,...,N\}}-\log(T'_{s_a, s_b})T_{s_as_b},
    \end{aligned}
\end{equation}
where $T'_{s_a, s_b}$ is the learned transition probability from state $s_a$ to state $s_b$ and  $T_{s_a, s_b}$ is the true transition probability.

\subsubsection{Results and Analysis}
We compare the proposed VRNN-LinearCRF against other unsupervised methods: K-means clustering, Hidden Markov Model, D-VRNN \cite{shi-etal-2019-unsupervised} and VRNN with vanilla attention. D-VRNN is similar to our work but without structured attention. We use a bidirectional LSTM with 300 hidden units as the sentence encoder and a forward LSTM for decoding. 300-dimensional word embeddings are initialized with GloVe word embedding \cite{pennington-etal-2014-glove}. A dropout rate of 0.5 is adopted during training. We set the BOW-loss weight $\lambda$ to be 0.5. The whole network is trained with the Adam optimizer with a learning rate of 0.001 on GTX Titan X GPUs for 60 epochs. The training takes on average 11.2 hours to finish. 

\begin{figure}[ht]
\vskip 0.2in
\begin{center}
\centerline{\includegraphics[width=\linewidth]{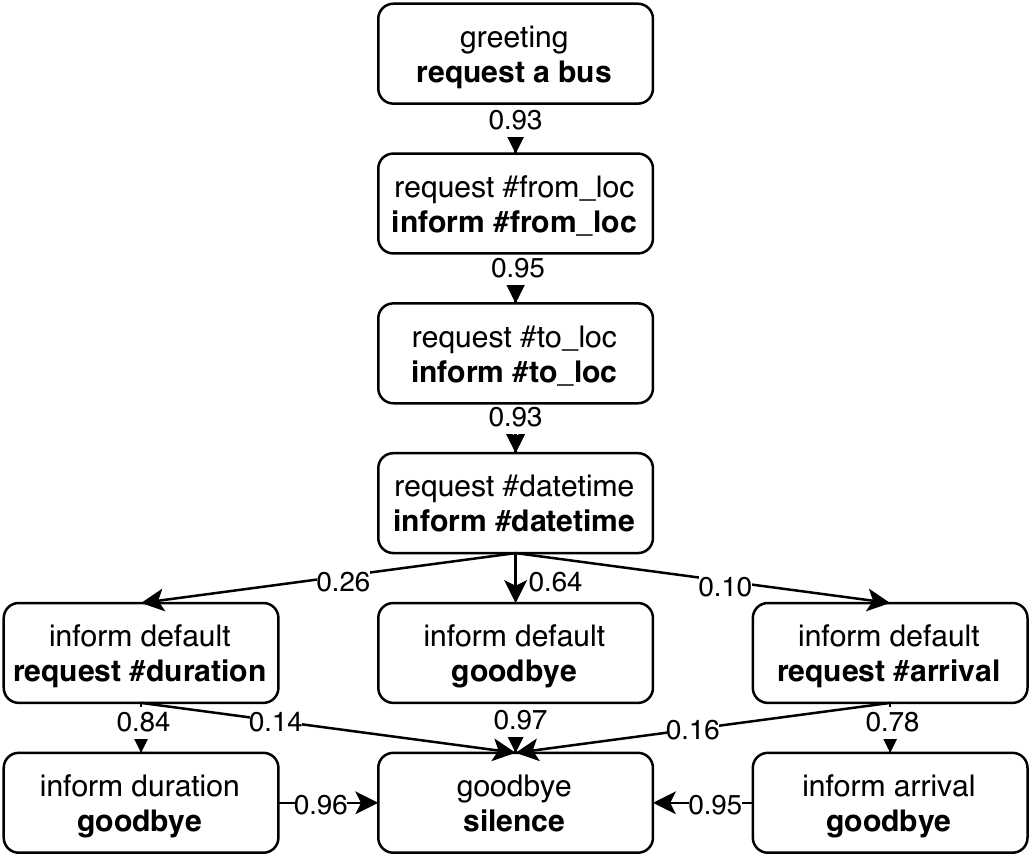}}
\caption{Learned semantic structure of SimDial bus domain \cite{zhao-eskenazi-2018-zero}. User intents are marked in bold. Transitions with $P < 0.1$ are omitted.}
\label{fig:learned_bus}
\end{center}
\vskip -0.2in
\end{figure}
\begin{figure}[ht]
\begin{center}
\centerline{\includegraphics[width=\linewidth]{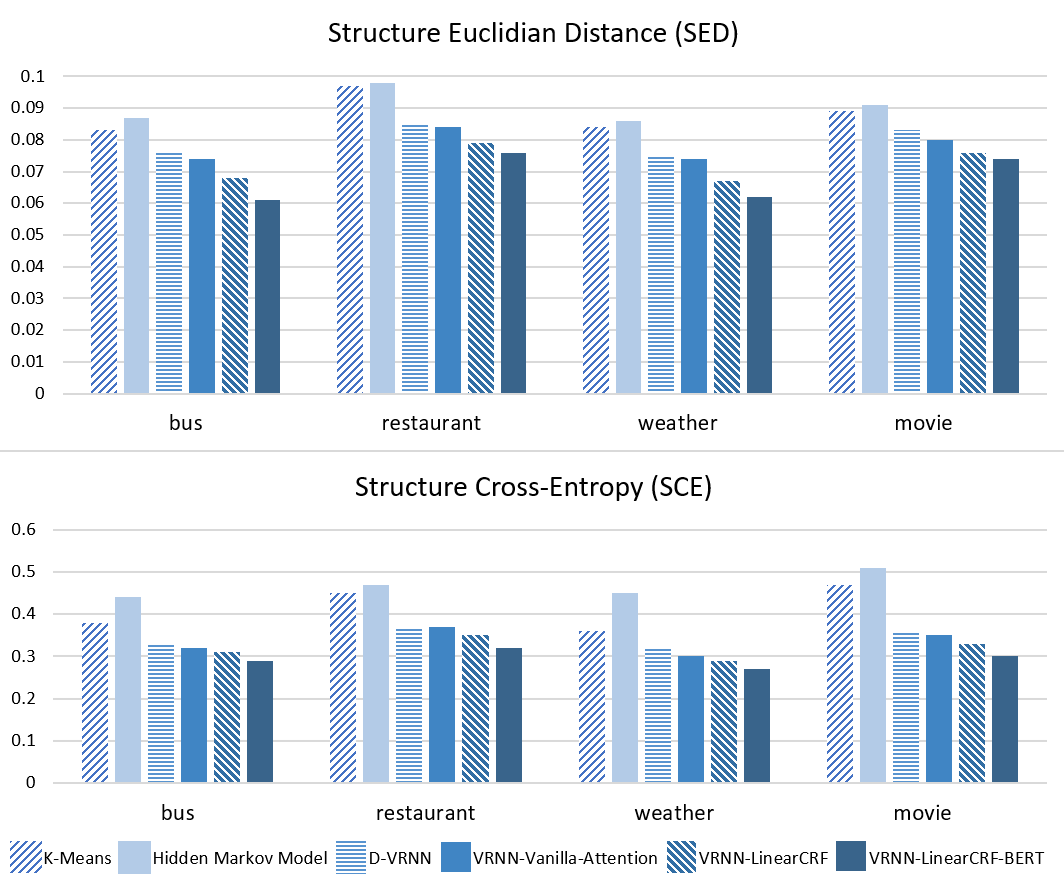}}
\caption{All models' performance in (a) Structure Euclidean Distance (SED) and (b) Structure Cross-Entropy (SCE) in four dialogue domains.}
\label{fig:d1_d2}
\end{center}
\end{figure}

To evaluate the learned structure, we compare VRNN-LinearCRF's output in Figure \ref{fig:learned_bus} with the ground truth dialogue structure in Figure \ref{fig:bus}.  A dialogue structure learned by VRNN without structured attention is also shown in the Appendix. We find our method generates similar structure compared to ground truth in the bus domain. Figure \ref{fig:d1_d2} shows all models' quantitative results. Having a lower value in SED and SCE indicates the learned structure is closer to the ground truth and better. Our method with BERT, VRNN-LinearCRF-BERT performs the best. K-means clustering performs worse than VRNN-based models because it only considers utterances' surface format and ignores the context information. Hidden Markov Model is similar to VRNN but lacks a continuous propagating hidden state layer. VRNN-LinearCRF observes the entire history of latent states but ignores the redundant transitions due to the structure attention. The model's performance further improves when replacing the vanilla LSTM encoder with a large scale pre-trained encoder like BERT \cite{devlin-etal-2019-bert}, as BERT provides better representations.

\subsection{Interactive Structure Learning in Multi-party Dialogues}
\begin{table*}[hbt!]
\centering
\begin{tabular}{|l|llll|l|l|}
\hline
Model                   & BLEU1 & BLEU2 & BLEU3 & BLEU4 & METEOR & ROUGE\textsubscript{L} \\ \hline
HRED                    & 10.54       & 4.63       & 2.67       & 1.53       & 4.22        & 10.14       \\
GSN No-speaker (1-iter) & 9.23        & 3.32       & 1.89       & 1.24       & 3.57        & 8.12       \\
GSN No-speaker (2-iter) & 11.32       & 4.89       & 2.94       & 1.54       & 4.12        & 10.15       \\
GSN No-speaker (3-iter) & 11.42       & 4.81       & 3.11       & 1.87       & 4.51        & 10.29       \\ \hline
GSN W-speaker (1-iter)  & 10.11       & 3.75       & 1.93       & 1.31       & 3.56        & 9.89       \\
GSN W-speaker (2-iter)  & 11.43       & 4.90       & 2.99       & 1.63       & 4.32        & 10.34       \\
GSN W-speaker (3-iter)  & \textbf{11.52}       & \textbf{4.93}       & 3.23       & 1.91       & \textbf{4.77}        & \textbf{11.21}       \\ \hline
VRNN-Dependency-Tree                    & 11.23       & 4.92       & \textbf{3.24}       & \textbf{1.92}       & 4.69        & 10.88       \\ \hline
\end{tabular}
\caption{Different methods' experiment results on Ubuntu dataset.}
\label{tab:automated-result}
\end{table*}
We extend our method to learn interactive structure in multi-party dialogues. Specifically, we detect each utterance's speaker and addressee by constructing an utterance dependency tree.

\subsubsection{Datasets}
We use Ubuntu Chat Corpus \cite{uthus2013ubuntu} as the dataset to study interactive structure since it provides the ground-truth of speaker/addressee information for evaluation. Though every record of Ubuntu Chat Corpus contains clear speaker ID, only part of the data has implicit addressee ID, coming as the first word in the utterance. We select addressee ID that appeared in a limited context and extract dialogue sessions with all utterances having verified speaker ID and addressee ID. We extract 20k dialogues with length ranging from 7 to 8 turns. Table \ref{tab:ubuntu-example} shows an example dialogue.
\begin{table}[]
\small
\begin{tabular*}{\linewidth}{ccl}
\hline
From    &To     &Utterance                      \\ \hline
$p_1$	&$p_2$	&I know upgrading always got hardon\\
        &       &settings to new system..\\
$p_3$	&$-$    &And the description of the settings is even\\
        &       &wrong\\
$p_1$	&$p_2$	&So these days i always clean install\\
$p_2$	&$p_1$	&Yeah, i think i will end up doing it\\
$p_2$	&$p_1$	&Do you happen to know if 12.10 install\\
        &       &will let me install grub2 to partition instead\\
        &       &of mbr without any extra tweaks?\\
$p_1$	&$p_2$	&I think default clean install will install\\
        &       &grub2 on first section of your hd\\
$p_4$	&$p_2$	&No\\ \hline
\end{tabular*}
\caption{Multi-party dialogue example in Ubuntu Chat Corpus \cite{uthus2013ubuntu}.}
\label{tab:ubuntu-example}
\end{table}

\subsubsection{Results and Analysis}
Considering Ubuntu Chat Corpus have a large number of technical terminologies, we use a relatively larger vocabulary size of 30k. We use LSTMs and BERT as the sentence embedding encoder and two GRU \cite{chung2014empirical} layers with 300 hidden units each as the decoder. The model converges after 100 epochs on GTX Titan X GPUs. The training procedure takes about 54 hours.

To evaluate the learned utterance dependency tree, we compare it with the annotated speaker-addressee relation and find 68.5\% utterances are assigned the correct parents. This is a reasonable number because the dependency relationship does not fully rely on the speaker/addressee information in a chatroom. A different interlocutor could answer others' questions even when the questions were not addressed to him/her. Figure \ref{fig:visualization} visualizes the learned interactive structure from the example in Table \ref{tab:ubuntu-example}. Specifically, utterance 4 largely depends on utterance 3, while utterance 6 and 7 are answering the question from utterance 5.

We also compare the model's generation performance with \textit{Hierarchical Recurrent Encoder-Decoder} (HRED) and \textit{Graph-Structured Network} (GSN) \cite{hu2019gsn}. The GSN model uses the annotated speaker/addressee information to construct a dialogue graph for utterance encoding iteration. However, this is not required by our VRNN-Dependency-Tree since we generate the original dialogues while learning a dependency structure. For consistent comparison with previous work, we evaluate all models with BLEU 1 to 4, METEOR, and ROUGE\textsubscript{L} with the package in \cite{chen2015microsoft}. All results are shown in Table \ref{tab:automated-result}. We observe that the proposed VRNN-Dependency-Tree model without using any speaker annotation achieves similar generation performance compared to the state-of-the-art method, GSN with speaker annotation. 

\section{Conclusion}
This paper proposed to inject structured attention into variational recurrent neural network models for unsupervised dialogue structure learning. We explored two different structure inductive biases: linear CRF for utterance-level semantic structure induction in two-party dialogues; and non-projective dependency tree for interactive structure learning in multi-party dialogues. Both models are proved to have a better structure learning performance over the state of the art algorithms. In the future, we will further explore how to explicitly incorporate linguistics information, such as named entities into the latent states.

\section*{Acknowledgments}
We would like to thank Ziheng Xu for providing scripts to process dialogue datasets. We also thank the outstanding reviewers for their helpful comments to improve our manuscript.

\bibliographystyle{acl_natbib}
\bibliography{emnlp2020}

\appendix
\section{Appendices}
\label{sec:appendix}
\subsection{Learned Structures of SimDial}
\begin{figure}[ht]
\vskip 0.2in
\begin{center}
\centerline{\includegraphics[width=\linewidth]{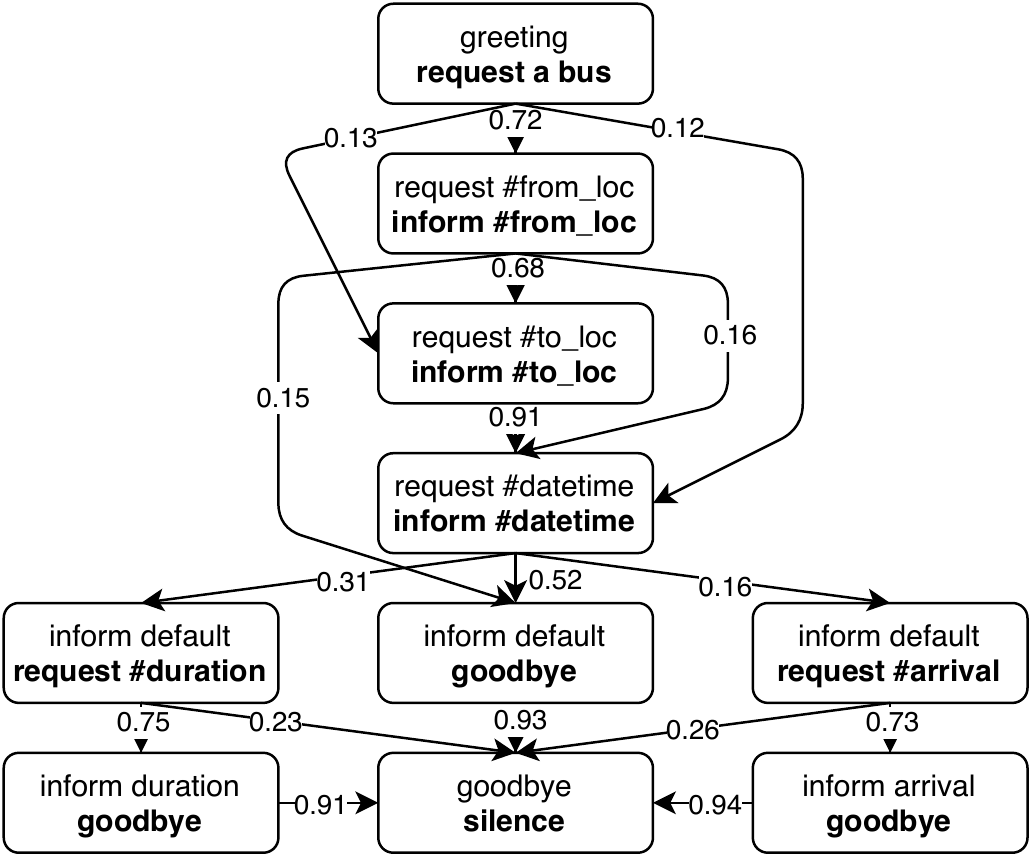}}
\caption{Learned dialogue structure from VRNN without structured attention in SimDial bus domain.}
\label{fig:learned_bus_vrnn}
\end{center}
\vskip -0.2in
\end{figure}

\subsection{Inside-Outside Algorithm}
\begin{algorithm*}[h]
  \caption{Inside-Outside for Non-projective Dependency Tree Attention}
  \label{alg:inside-outside}
\begin{algorithmic}
  \STATE {\bfseries Input:} potential $\theta_{ij}$
  \STATE $\alpha,\beta \leftarrow -\infty$
  \FOR{$i=1,...,n$}
    \STATE $\alpha[i,i,L,1] \leftarrow 0$
    \STATE $\alpha[i,i,R,1] \leftarrow 0$
  \ENDFOR
  \STATE $\beta[1,n,R,1] \leftarrow 0$
  \FOR{$k=1,...,n$}
    \FOR{$s=1,...,n-k$}
        \STATE $t \leftarrow s+k$
        \STATE $\alpha[s,t,R,0] \leftarrow \bigoplus_{u \in [s,t-1]}\alpha[s,u,R,1]\otimes\alpha[u+1,t,L,1]\otimes\theta_{st}$
        \STATE $\alpha[s,t,L,0] \leftarrow \bigoplus_{u \in [s,t-1]}\alpha[s,u,R,1]\otimes\alpha[u+1,t,L,1]\otimes\theta_{ts}$
        \STATE $\alpha[s,t,R,1] \leftarrow \bigoplus_{u \in [s+1,t]}\alpha[s,u,R,0]\otimes\alpha[u,t,R,1]$
        \STATE $\alpha[s,t,L,1] \leftarrow \bigoplus_{u \in [s,t-1]}\alpha[s,u,L,1]\otimes\alpha[u,t,L,0]$
    \ENDFOR
  \ENDFOR
  \FOR{$k=n,...,1$}
    \FOR{$s=1,...,n-k$}
        \STATE $t \leftarrow s+k$
        \FOR{$u=s+1,...,t$}
            \STATE $\beta[s,u,R,0] \leftarrow_\oplus \beta[s,t,R,1] \otimes \alpha[u,t,R,1]$
            \STATE $\beta[u,t,R,1] \leftarrow_\oplus \beta[s,t,R,1] \otimes \alpha[s,u,R,0]$
        \ENDFOR
        \IF{$s>1$}
            \FOR{$u=s,...,t-1$}
                \STATE $\beta[s,u,L,1]\leftarrow_\oplus \beta[s,t,L,1] \otimes \alpha[u,t,L,0]$
                \STATE $\beta[u,t,L,0]\leftarrow_\oplus \beta[s,t,L,1] \otimes \alpha[s,u,L,1]$
            \ENDFOR
        \ENDIF
        \FOR{$u=s,...,t-1$}
            \STATE $\beta[s,u,R,1]\leftarrow_\oplus \beta[s,t,R,0] \otimes \alpha[u+1,t,L,1]\otimes\theta_{st}$
            \STATE $\beta[u+1,t,L,1]\leftarrow_\oplus \beta[s,t,R,0] \otimes \alpha[s,u,R,1]\otimes\theta_{st}$
        \ENDFOR
        \IF{$s>1$}
            \FOR{$u=s,...,t-1$}
                \STATE $\beta[s,u,R,1]\leftarrow_\oplus \beta[s,t,L,0] \otimes \alpha[u+1,t,L,1]\otimes\theta_{ts}$
                \STATE $\beta[u+1,t,L,1]\leftarrow_\oplus \beta[s,t,L,0] \otimes \alpha[s,u,R,1]\otimes\theta_{ts}$
            \ENDFOR
        \ENDIF
    \ENDFOR
  \ENDFOR
  \STATE $A \leftarrow \alpha[1,n,R,1]$
  \FOR{$s=1,...,n-1$}
    \FOR{$t=s+1,...,n$}
        \STATE $p[s,t] \leftarrow exp(\alpha[s,t,R,0]\otimes\beta[s,t,R,0]\otimes-A)$
        \IF{$s>1$}
            \STATE $p[t,s] \leftarrow exp(\alpha[s,t,L,0]\otimes\beta[s,t,L,0]\otimes-A)$
        \ENDIF
    \ENDFOR
  \ENDFOR
\end{algorithmic}
\end{algorithm*}




\end{document}